\documentclass[letterpaper, 10 pt, conference]{ieeeconf}  

\IEEEoverridecommandlockouts                              

\usepackage[T1]{fontenc}
\usepackage[utf8]{inputenc}
\usepackage{authblk}
\usepackage{graphicx}
\usepackage{array}
\usepackage{float}
\usepackage{float}
\usepackage{booktabs}
\usepackage{bbding}
\usepackage{etoolbox}
\usepackage{soul}
\usepackage{siunitx} 
\usepackage{algorithm}
\usepackage{algorithmic}
\usepackage{cite}
\usepackage{balance}
\usepackage{amsmath, amssymb}
\usepackage{makecell}
\usepackage{multirow}
\usepackage{hyperref}

\usepackage{cite}
\usepackage{amsmath,amssymb,amsfonts}
\usepackage{algorithmic}
\usepackage{graphicx}
\usepackage{textcomp}

\usepackage{multirow}
\usepackage{adjustbox}
\usepackage{booktabs}
\usepackage{url}
\usepackage{xcolor}
\usepackage{color}

\begin{document}
\title{\LARGE \bf
PitVQA++: Vector Matrix-Low-Rank Adaptation for Open-Ended Visual Question Answering in Pituitary Surgery}
\author{Runlong He, Danyal Z. Khan, Evangelos B. Mazomenos, Hani J. Marcus, Danail Stoyanov, Matthew J.\\ Clarkson, and Mobarakol Islam
\thanks{This work was supported in whole, or in part, by the Engineering and Physical Sciences Research Council (EPSRC) [EP/W00805X/1, EP/Z534754/1].}
\thanks{Runlong He, Evangelos Mazomenos, Matthew J. Clarkson, and Mobarakol Islam are with the UCL Hawkes Institute and the Department of Medical Physics \& Biomedical Engineering, University College London, UK (E-mail: runlong.he.23@ucl.ac.uk; e.mazomenos@ucl.ac.uk; m.clarkson@ucl.ac.uk; mobarakol.islam@ucl.ac.uk).}
\thanks{Danail Stoyanov is with the UCL Hawkes Institute and the Department of Computer Science, University College London, UK (E-mail: danail.stoyanov@ucl.ac.uk).}
\thanks{Danyal Z. Khan, and Hani J. Marcus are with the UCL Hawkes Institute and the Department of Neurosurgery, National Hospital for Neurology and Neurosurgery, UK (E-mail: d.khan@ucl.ac.uk; h.marcus@ucl.ac.uk).}
}
\maketitle
\thispagestyle{empty}
\pagestyle{empty}

\begin{abstract}
Vision-Language Models (VLMs) in visual question answering (VQA) offer a unique opportunity to enhance intra-operative decision-making, promote intuitive interactions, and significantly advancing surgical education. However, the development of VLMs for surgical VQA is challenging due to limited datasets and the risk of overfitting and catastrophic forgetting during full fine-tuning of pretrained weights. While parameter-efficient techniques like Low-Rank Adaptation (LoRA) and Matrix of Rank Adaptation (MoRA) address adaptation challenges, their uniform parameter distribution overlooks the feature hierarchy in deep networks, where earlier layers, that learn general features, require more parameters than later ones. This work introduces PitVQA++ with an open-ended PitVQA dataset and vector matrix-low-rank adaptation (Vector-MoLoRA), an innovative VLM fine-tuning approach for adapting GPT-2 to pituitary surgery. Open-Ended PitVQA comprises around 101,803 frames from 25 procedural videos with 745,972 question-answer sentence pairs, covering key surgical elements such as phase and step recognition, context understanding, tool detection, localization, and interactions recognition. Vector-MoLoRA incorporates the principles of LoRA and MoRA to develop a matrix-low-rank adaptation strategy that employs vector ranking to allocate more parameters to earlier layers, gradually reducing them in the later layers. Our approach, validated on the Open-Ended PitVQA and EndoVis18-VQA datasets, effectively mitigates catastrophic forgetting while significantly enhancing performance over recent baselines. Furthermore, our risk-coverage analysis highlights its enhanced reliability and trustworthiness in handling uncertain predictions. Our source code and dataset is available at~\url{https://github.com/HRL-Mike/PitVQA-Plus}.
\end{abstract}

\section{INTRODUCTION}
\label{sec:introduction}
The integration of Vision-Language Models (VLMs) into surgical workflows has the potential to revolutionize intra-operative decision-making, fostering intuitive surgeon-AI interactions and transforming surgical education. In the context of visual question answering (VQA), VLMs can provide real-time insights into surgical phases, steps, and instrument usage, addressing the growing demand for intelligent, context-aware systems in operating rooms. For complex procedures such as endonasal pituitary surgery, which demands precise anatomical navigation and dynamic decision-making \cite{khan2023current}, VLMs could offer invaluable support by analyzing intraoperative videos and images to predict surgical outcomes and guide interventions \cite{he2024pitvqa, moghani2024sufia}. However, developing VLMs tailored for surgical applications presents unique challenges, primarily due to the scarcity of domain-specific datasets and the increased risk of overfitting during fine-tuning of pretrained models originally trained on massive datasets, which can compromise their generalization capabilities \cite{li2024llava, balne2024parameter}. Existing parameter-efficient fine-tuning techniques, such as Low-Rank Adaptation (LoRA) \cite{hu2022lora} and Matrix of Rank Adaptation (MoRA) \cite{jiang2024mora}, offer promising solutions but fail to account for the hierarchical feature distribution inherent in deep neural networks. This shortcoming motivates the need for innovative approaches that adaptively allocate parameters to layers based on their roles in feature extraction and representation.

\begin{figure}[t]
    \centering
    \includegraphics[width=\columnwidth]
    {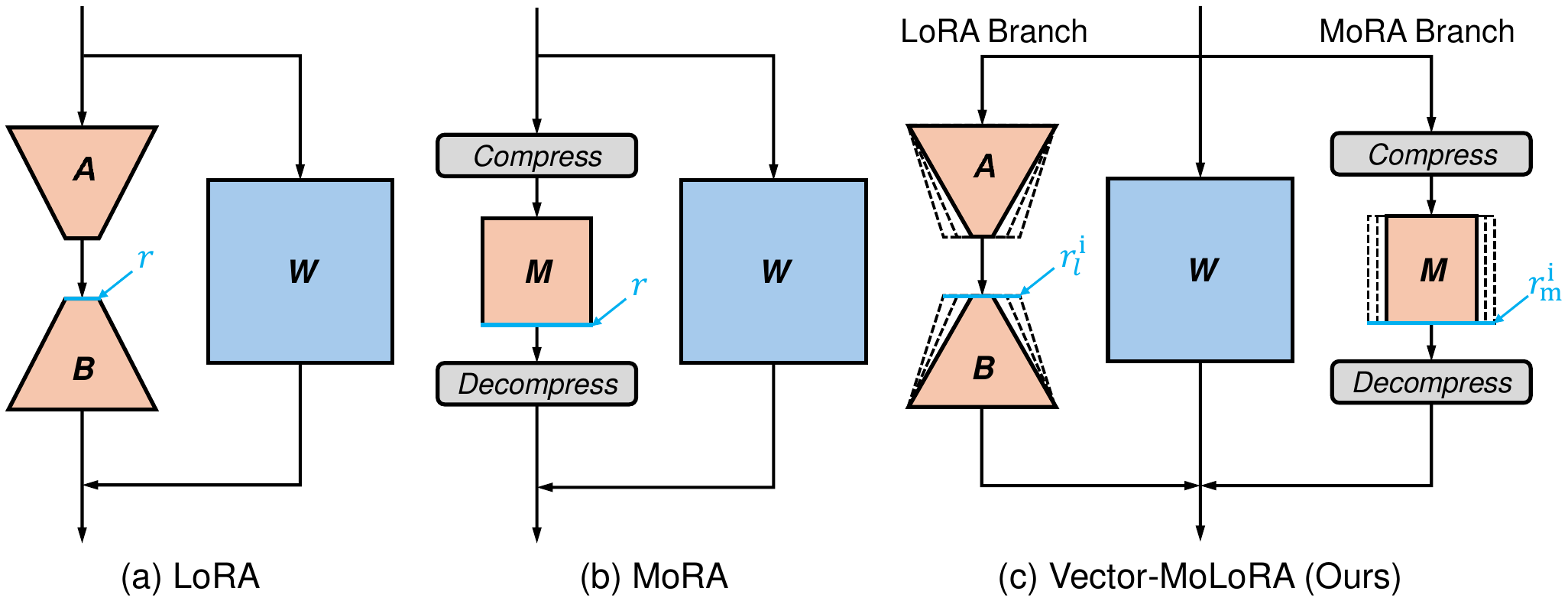}
    \caption{LoRA \cite{hu2022lora}, MoRA \cite{jiang2024mora}, and the proposed Vector-MoLoRA serve as examples of parameter-efficient fine-tuning methods, with Vector-MoLoRA leveraging a variable rank vector to align with the hierarchical structure of neural networks and enhance fine-tuning performance.}
    \label{fig:lora_mora_molora}
\end{figure}

Bridging this gap necessitates the creation of enriched surgical VQA datasets that challenge VLMs with tasks demanding deeper reasoning and diverse contextual interpretations. Several surgical VQA datasets have been proposed for different surgical scenarios, including EndoVis18-VQA \cite{seenivasan2022surgical} for nephrectomy, SSG-VQA \cite{yuan2024advancing} for laparoscopic cholecystectomy, and Kvasir-VQA \cite{gautam2024kvasir} for colonoscopic procedures. Nevertheless, most existing surgical VQA datasets are designed for basic tasks or simple reasoning tasks, such as instrument recognition \cite{yuan2024advancing}, instrument localization \cite{gautam2024kvasir}, and tool-tissue interactions \cite{seenivasan2022surgical}. Additionally, these datasets are primarily close-ended, focused on classification tasks with single-word answers, limiting VLM's natural language generative capability and confining it to discriminative outcomes. Such constrained answering scheme, containing very limited contextual information, also impedes surgeon-AI interaction. More importantly, there is no open-ended dataset specifically designed for endonasal pituitary surgery, an intricate and specialized procedure that requires real-time support for both intervention and education.

Previous studies of SSG-VQA \cite{yuan2024advancing}, SurgicalGPT \cite{seenivasan2023surgicalgpt}, and SurgicalVQA \cite{seenivasan2022surgical} primarily obtained surgical VQA systems by fully fine-tuning existing VQA models on surgical data. Due to limited data availability, this approach can lead to overfitting and catastrophic forgetting \cite{li2024llava}. Potential solutions include parameter-efficient fine-tuning (PEFT) techniques, such as LoRA \cite{hu2022lora}, MoRA \cite{jiang2024mora}, ALoRA \cite{liu2024alora}, QLoRA \cite{dettmers2023qlora}, and AdaLoRA \cite{zhang2023adalora}. These PEFT methods mitigate overfitting and catastrophic forgetting by storing domain-specific knowledge in lightweight adapters during fine-tuning, while keeping the backbone network's pre-trained weights frozen. While PEFT techniques demonstrate promising performance on small datasets, their uniform parameter distribution and structure design fail to consider the inherent feature hierarchy in deep networks, where earlier layers learn general features that require more parameters compared to later layers \cite{xu2024correlating}. Therefore, there is an urgent need to develop efficient adaptation techniques that address catastrophic forgetting and optimize the learning dynamics of vision-language models for image-guided intervention.

In this paper, we develop an open-ended dataset and an innovative parameter-efficient fine-tuning technique for surgical VQA. Our dataset targets endonasal pituitary surgery and consists of comprehensive intraoperative descriptions, encompassing both real-time surgical information, including surgical phases, steps, instrument usage, activities, and predictive procedural planning. To address these fine-tuning challenges, we introduce Vector-MoLoRA, which follows the dynamic features of deep neural networks and employs variable rank vector-based adapters, as shown in Fig.\ref{fig:lora_mora_molora} (c), to enhance model performance.

Our contributions and findings are below:
\begin{enumerate}
    \item[--] We build a specialized open-ended VQA dataset for endonasal pituitary surgery, comprising 101,803 surgical frames along with 745,972 question-answer sentence pairs covering key surgical concepts.
    
    \item[--] We propose Vector-MoLoRA, an efficient adaptation technique that leverages the principles of LoRA and MoRA and incorporate the hierarchy feature of deep networks to mitigate catastrophic forgetting when fine-tuning foundational vision-language models. 

    \item[--] We evaluate our adaptation technique on both public benchmark and open-ended PitVQA datasets, demonstrating superior performance over state-of-the-art methods. Risk-coverage analysis reveals Vector-MoLoRA's enhanced reliability and trustworthiness, with fewer clinician referrals needed for uncertain samples.
\end{enumerate}

This work enhances our prior research \cite{he2024pitvqa} by introducing Vector-MoLoRA, a novel adaptation technique that mitigates catastrophic forgetting and optimizes fine-tuning performance. It expands the dataset with an open-ended PitVQA, enabling sentence-based question-answering and complex reasoning tasks. This work also adapts GPT-2 as a vision-language model for sentence generation instead of classification, incorporating dynamic parameter allocation for efficient learning. While prior work focused on classification tasks and comparisons with classification baselines, this work compares the performance with VQA sentence generation baselines and demonstrates the mitigation of overfitting and catastrophic forgetting challenges.

\section{RELATED WORK}

\subsection{Visual Question Answering}
Early VQA methods relied on basic networks and simple fusion mechanisms, while subsequent approaches enhanced cross-modal interaction through question-guided attention and region-based representations. Latest research focuses on a broad Vision-Language Pre-training (VLP) paradigm based on the Transformer architecture \cite{vaswani2017attention}. In this paradigm, approaches can be categorized into three types according to the ways of visual feature extraction: (I) Object-based methods (e.g., ViLBERT \cite{lu2019vilbert}, VL-BERT \cite{su2020vl}, and VisualBERT \cite{li2019visualbert}) that utilize Faster R-CNN for ROI feature extraction; (II) Convolution-based methods like Pixel-BERT \cite{huang2020pixel} that employ CNNs; and (III) Image-patch-based methods that divide images into sequences of patches for processing, such as CLIP \cite{radford2021learning}, BLIP \cite{li2022blip}, and LLaVA \cite{liu2023llava}. These models, trained on large-scale datasets to align visual and textual features into a shared embedding space, demonstrate strong few-shot transfer capabilities on downstream tasks such as VQA. Recent works \cite{zhang2023biomedclip, li2024llava} attempt to transfer general VQA models to the biomedical domain through post-training on domain datasets. These works require high computational cost and have not been considered for surgical applications.

\subsection{Surgical Visual Question Answering}
Recent studies mainly applied pre-trained visual and textual encoders on surgical VQA tasks~\cite{seenivasan2022surgical, seenivasan2023surgicalgpt, yuan2024advancing, he2024pitvqa, Du2024llm}. Specifically, the visual features of surgical frames are extracted by CNNs (e.g., VGGNet, ResNet \cite{he2016deep}) or Vision transformer (ViT) \cite{dosovitskiy2021image}, and the textual features of surgical questions are gained through stacked transformer layers \cite{li2019visualbert}. Early approaches, such as VisualBERT-RM \cite{seenivasan2022surgical} and SurgicalGPT \cite{seenivasan2023surgicalgpt}, relied on concatenating visual and textual representations, followed by a self-attention layer. However, this concatenation leads to computational inefficiency due to the elongated embeddings, often requiring additional MLP layers for feature projections~\cite{seenivasan2023surgicalgpt}. Yuan et al. \cite{yuan2024advancing} used scene graph generation for surgical VQA, but its multistage training is complex, computationally expensive, and reliant on prior task predictions. Existing methods primarily focus on close-ended surgical VQA, while there is a lack of datasets and approaches suitable for open-ended surgical VQA tasks.

With the rising interest in foundation models, researchers begin to apply large language models (LLMs) for surgical VQA tasks. He et al.~\cite{he2024pitvqa} employed cross-attention mechanism to model the correlations between visual and textual features, and a LLM to decode vision-language embeddings. However, fully fine-tuning foundation models on small domain datasets can lead to catastrophic forgetting, where the LLM may 'forget' its pre-trained general knowledge due to overfitting on limited data \cite{balne2024parameter, bafghi2025fine}. Du et al.~\cite{Du2024llm} proposed a multi-teacher continual learning framework that balances knowledge from a frozen LLM and a medical expert model. While this approach attempts to mitigate catastrophic forgetting, it requires a carefully designed student model. Furthermore, the need to balance multiple objectives, including distillation and task-specific losses, increases the training complexity.

\subsection{Parameter-efficient Fine-tuning}
Parameter-efficient Fine-tuning (PEFT) updates only a small subset of model parameters while keeping the majority frozen. LoRA\cite{hu2022lora} decomposes weight updates into low-rank matrices to enable efficient adaptation, but its low-rank updating mechanism may limit the model's ability to learn new knowledge effectively. MoRA\cite{jiang2024mora} addresses this limitation by employing a square matrix for matrix-rank updating while maintaining the same parameter count as LoRA, achieving better performance. DoRA\cite{jiang2024mora} decomposes pre-trained weights into magnitude and direction components, using LoRA for directional updates to bridge the performance gap with full fine-tuning while maintaining inference efficiency. ALoRA\cite{liu2024alora} tackles the uniform rank limitation in LoRA by adaptively allocating different rank sizes across layers based on parameter importance, achieving better parameter efficiency. QLoRA\cite{dettmers2024qlora} further reduces memory requirements by combining 4-bit quantization with LoRA, enabling fine-tuning of large models on consumer GPUs at the cost of increased training time. However, these methods do not explicitly consider the inherent hierarchical structure of deep neural networks, where earlier layers extract general features with larger parameter spaces while later layers focus on task-specific features with more compact representations \cite{xu2024correlating}.

\section{METHODOLOGY}
\subsection{Preliminaries}
\subsubsection{Low-Rank Adaptation (LoRA)}
LoRA is designed to minimize trainable parameters during fine-tuning by decomposing weight updates into two low-rank matrices. Specifically, it keeps the original pre-trained weights $W_0 \in \mathbb{R}^{d\times k}$ frozen and introduces matrices $B \in \mathbb{R}^{d\times r}$ and $A \in \mathbb{R}^{r\times k}$ for efficient adaptation, where the rank $r$ is significantly smaller than both dimensions $d$ and $k$. This approach leverages the concept of "intrinsic rank" to prevent catastrophic forgetting, with the final forward computation combining the original transformation $W_0x$ with the low-rank adaptation $BAx$, as shown in Equation \ref{eq:lora}:
\begin{equation}
h = W_0x + \Delta Wx = W_0x + BAx
\label{eq:lora}
\end{equation}

where both the pre-trained weights $W_0$ and the low-rank update $\Delta W = BA$ operate on the input $x \in \mathbb{R}^k$ with their outputs being summed. This adaptation mechanism can be applied to all dense layers in transformer blocks, including query (q), key (k), value (v), and output (o) projections.

\begin{figure*}[ht]
    \centering
    \includegraphics[width=1\textwidth]
    {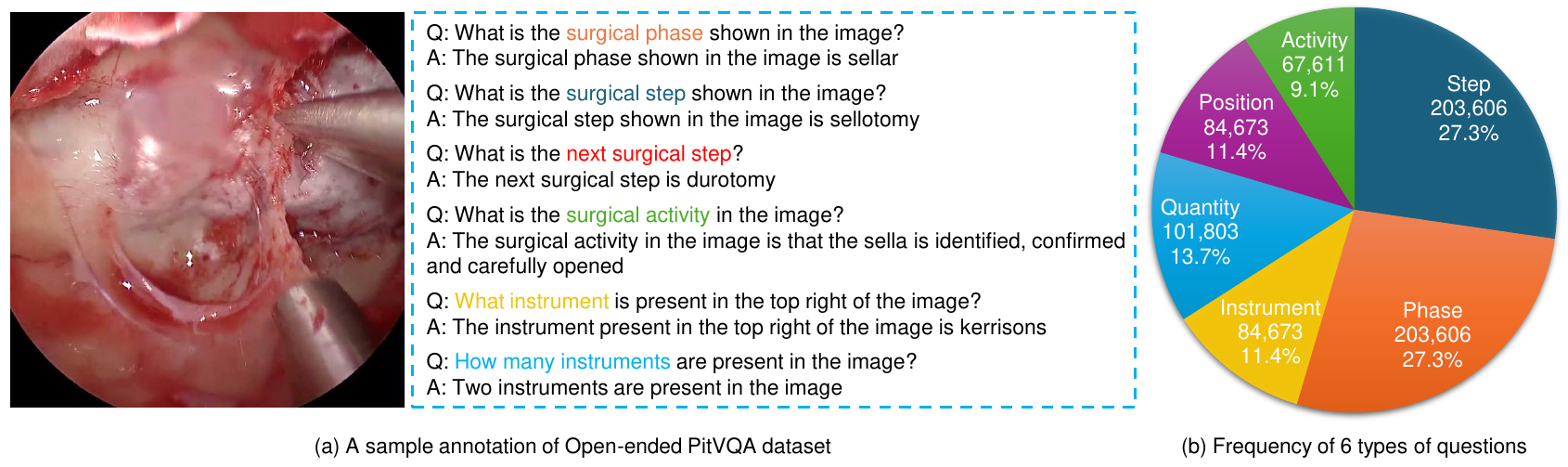}
    \caption{Open-ended PitVQA dataset of visual questions answering for pituitary surgery. There are overall 6 types of question-answer pairs covering surgical phases, steps, instruments, quantity, positions and activities.}
    \label{fig:pitvqa_dataset}
\end{figure*}

\subsubsection{MoRA}
MoRA aims to achieve matrix-rank updates while maintaining the same parameter efficiency as LoRA. Given a pre-trained weight matrix $W_0 \in \mathbb{R}^{d\times k}$, MoRA employs a square matrix $M \in \mathbb{R}^{\hat{r}\times\hat{r}}$ for updating:
\begin{equation}
h = W_0x + \Delta Wx = W_0x + f(Mg(x))
\end{equation}

where $g(\cdot)$ and $f(\cdot)$ are non-parametric operators for compression and decompression, respectively, and $M$ is a trainable square matrix that processes the compressed representation. Specifically, $g: \mathbb{R}^k \rightarrow \mathbb{R}^{\hat{r}}$ reduces the $k$-dimensional input to $\hat{r}$ dimensions, $f: \mathbb{R}^{\hat{r}} \rightarrow \mathbb{R}^d$ increases the $\hat{r}$ dimensions back to $d$ dimensions, while $M: \mathbb{R}^{\hat{r}} \rightarrow \mathbb{R}^{\hat{r}}$ transforms within the compressed space.
These two operators also have corresponding function, $\overline{g}: \mathbb{R}^{\hat{r} \times \hat{r}} \to \mathbb{R}^{\hat{r} \times k}$ and $\overline{f}: \mathbb{R}^{\hat{r} \times k} \to \mathbb{R}^{d\times k}$, to transform $M$ into $\Delta W$. 
For any input $x \in \mathbb{R}^{k}$, the following should hold:
\begin{equation}
f(Mg(x)) = \Delta W x
\end{equation}

where $\Delta W = \overline{f}(\overline{g}(M))$.
Following RoPE\cite{su2024roformer}, the compression and decompression operators can be expressed as:
\begin{align}
g(x) &= [b^0\:b^1\:\cdots\:b^{n-1}] \label{eq:g_func} \\
f(x) &= \text{concat}(x) \\
\Delta W &= \begin{bmatrix} 
P^0 & 0 & \cdots & 0 \\
0 & P^1 & \cdots & 0 \\
\vdots & \vdots & \ddots & \vdots \\
0 & 0 & \cdots & P^{n-1}
\end{bmatrix}
\end{align}

where \text{concat}(x) denotes the operation of flattening $x$ into a vector, $b^j$ and $P^j$, $j \in \{0, 1, \dots, n-1\}$, denote the  corresponding values of $x_{j\hat{r}:(j+1)\hat{r}}$ and $M$ after rotation, respectively. $b^j$ and $p^j$ are defined as follow:
\begin{align}
b^j &= \begin{bmatrix}
R_{\theta_{1,j}} & 0 & \cdots & 0 \\
0 & R_{\theta_{2,j}} & \cdots & 0 \\
\vdots & \vdots & \ddots & \vdots \\
0 & 0 & \cdots & R_{\theta_{\frac{\hat{r}}{2}},j}
\end{bmatrix} x_{j\hat{r}:(j+1)\hat{r}} \\
P^j &= M\begin{bmatrix}
R_{\theta_{1,j}} & 0 & \cdots & 0 \\
0 & R_{\theta_{2,j}} & \cdots & 0 \\
\vdots & \vdots & \ddots & \vdots \\
0 & 0 & \cdots & R_{\theta_{\frac{\hat{r}}{2}},j}
\end{bmatrix}
\end{align}

where $\theta_k = 10000^{-2(k-1)/\hat{r}}$, $k \in \{1, 2, \dots, \frac{\hat{r}}{2}\}$ and $R_{\theta_k,j} $ is a rotation matrix:
\begin{equation}\label{eq:rotation-mat}
R_{\theta_k,j} = \begin{bmatrix}
\cos j\theta_k & -\sin j\theta_k \\
\sin j\theta_k & \cos j\theta_k
\end{bmatrix}
\end{equation}

\subsubsection{Vector-LoRA}
Vector-LoRA\cite{zeinoddin2024dares} extends the traditional LoRA approach by introducing adaptive rank allocation across network layers. While maintaining the core low-rank adaptation principle, it uniquely assigns varying ranks to different layers based on their hierarchical importance in the network. The method defines rank adjustments through a rank vector $r_l$, as shown in equation \ref{eq:lora_rank_1}. For any adaptation layer $i$, the computation of Vector-LoRA is given by equation \ref{eq:vec_lora}:
\begin{equation}\label{eq:lora_rank_1}
\mathbf{r}_l = [r_1, r_2, r_3, ..., r_i]
\end{equation}
\begin{equation}\label{eq:vec_lora}
h(r_i) = W_ix_i + \Delta W_ix_i = W_ix_i + B_{r_i}A_{r_i}x_i
\end{equation}

where the input $x_i \in \mathbb{R}^{k}$ is processed by both the pre-trained weights $W_i$ and the low-rank update $\Delta W_i = B_{r_i}A_{r_i}$. $B_{r_i} \in \mathbb{R}^{d\times r_i}$ and $A_{r_i} \in \mathbb{R}^{r_i\times k}$ are fully connected layers used for compression and decompression, respectively. 
This approach allocates higher ranks to earlier layers where general feature learning occurs, progressively decreasing through later layers that focus on task-specific refinements, thereby optimizing both parameter efficiency and adaptation effectiveness.

\subsection{Proposed Method}
\subsubsection{Open-Ended PitVQA Dataset}
Our Open-ended PitVQA dataset consists of 25 videos of endoscopic pituitary surgery collected at the National Hospital of Neurology and Neurosurgery in London, UK. The videos are similar to the dataset used in the MICCAI PitVis challenge \cite{das2024pitvis}. The surgeries were recorded using a high-definition Karl Storz endoscope at 720p resolution and stored in .mp4 format. The study proceeded with patient informed consent and local governance committee approval. Following a standardized annotation framework derived from an international consensus study on pituitary surgery workflow \cite{marcus2021pituitary}, the videos were comprehensively annotated to include surgical phases, procedural steps, instrument utilization, and description of surgical activities. The annotation process involved collaborative work by two neurosurgical residents experienced in pituitary surgery, with final validation provided by an attending neurosurgeon.

We extracted surgical frames from each video at a rate of 1 fps and removed any frames that exhibited blur or occlusion. Our final dataset comprises 101,803 frames, with individual video contributions ranging from 2,443 to 7,179 frames. For each frame, we generated question-answer pairs using pre-defined templates, resulting in a collection of 745,972 pairs. The dataset contains 59 distinct questions distributed among 6 major categories, with lengths varying from 7 to 11 words. 
Fig.\ref{fig:pitvqa_dataset}(a) illustrates a representative frame with its corresponding question-answer pairs, while Fig.\ref{fig:pitvqa_dataset}(b) presents the frequency distribution of questions by category, revealing surgical activities and phases as the least and most frequent questions at 9.1\% and 27.3\% respectively.

\begin{figure*}[!ht]
    \centering
    \includegraphics[width=1\textwidth]
    {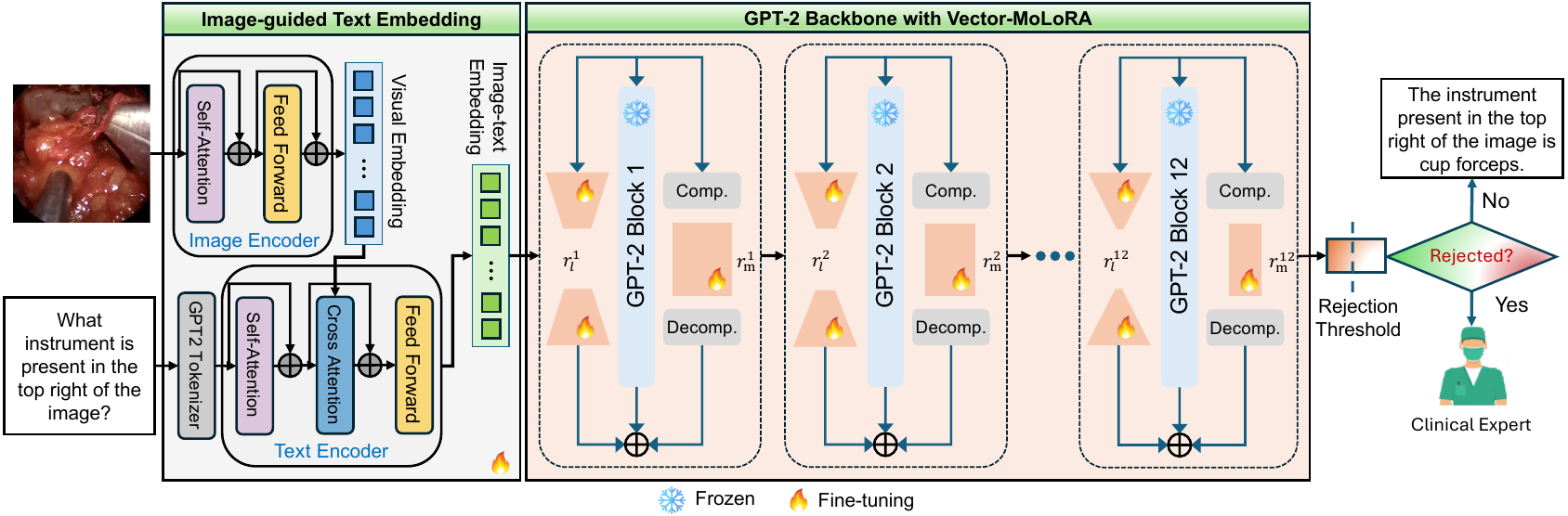}
    \caption{PitVQA++: The network consists of Image-grounded Text Embedding and GPT-2 Backbone with Vector-MoLoRA. The Vector-MoLoRA leverages two rank vectors to control the size of adapters in each layer. An entropy-based uncertainty threshold is incorporated to reject uncertain predictions and refer them to a clinician during inference, ensuring improved reliability in surgical decision support.}
    \label{fig:model_archi}
\end{figure*}

\subsubsection{Vector-MoLoRA}
The low-rank update matrix of LoRA or Vector-LoRA limits the ability to effectively learn new knowledge compared to fully fine-tuning (FFT) \cite{jiang2024mora}. While MoRA attempts to address this limitation by employing a square matrix to achieve high-rank updates, it uses a static rank allocation designed consistently across different layers. This fixed-rank design overlooks the hierarchical feature of neural networks, limiting its flexibility during fine-tuning. To address the limitations of low-rank updating and MoRA's rigid rank allocation, we design Vector-MoLoRA, which combines matrix-rank and low-rank adaptation with vector ranking for the surgical VQA task.

Vector-MoLoRA specifies rank adjustment through two separate rank vectors: $r_l$ defines the layer-wise LoRA rank, while $r_m$ determines the rank of MoRA matrices for each layer, as shown in Eq. \ref{eq:lora_vector_2} and \ref{eq:mora_vector}, respectively. 
\begin{align}
\mathbf{r}_l &= \begin{bmatrix}
r_l^1 & r_l^2 & \cdots & r_l^i
\end{bmatrix} \label{eq:lora_vector_2} \\
\mathbf{r}_m &= \begin{bmatrix}
r_m^1 & r_m^2 & \cdots & r_m^i
\end{bmatrix} \label{eq:mora_vector}
\end{align}

where $r_l^i$ denotes the LoRA rank at layer $i$, and $r_m^i$ indicates the dimension of MoRA matrices at the corresponding layer. The fundamental computation of Vector-MoLoRA is similar to Eq. \ref{eq:vec_lora} but extends it with layer-specific MoRA adaptation. The weight update consists of two components: $\Delta W_l$ for LoRA adaptation and $\Delta W_m$ for MoRA adaptation:
\begin{equation}\label{eq:vec-molora-1}
h = W_0x + \Delta W_lx + \Delta W_mx
\end{equation}

As shown in Fig. \ref{fig:model_archi}, for the transformer layer $i$ in the GPT-2 backbone, the updating of Vector-MoLoRA weights can be described by Eq. \ref{eq:vec_molora_2}: 
\begin{equation}\label{eq:vec_molora_2}
h(r_l^i, r_m^i) = W_ix_i + B_{r_l^i}A_{r_l^i}x_i 
                  + f_{r_m^i}(M_{r_m^i}g_{r_m^i}(x_i))
\end{equation}

where $W_i \in \mathbb{R}^{d\times k}$ is the original pre-trained weights, $B_{r_l^i} \in \mathbb{R}^{d\times r_l^i}$ and $A_{r_l^i} \in \mathbb{R}^{r_l^i\times k}$ are the LoRA adaptation matrices. $M_{r_m^i} \in \mathbb{R}^{r_m^i \times r_m^i}$ is a trainable MoRA square matrix. Functions $g_{r_m^i}(\cdot): \mathbb{R}^k \rightarrow \mathbb{R}^{r_m^i}$ and $f_{r_m^i}(\cdot): \mathbb{R}^{r_m^i} \rightarrow \mathbb{R}^d$ are non-parameterized compression and decompression operators that reduce the input dimension from $k$ to $r_m^i$ and expand the output dimension from $r_m^i$ to $d$, respectively. 

\subsubsection{PitVQA++ Network}
Our PitVQA++ network consists of an Image-grounded Text Embedding and a GPT-2 \cite{radford2019language} Backbone with Vector-MoLoRA, as shown in Fig. \ref{fig:model_archi}. These two modules are detailed below: 

\textit{Image-grounded Text Embedding:}
The Image-grounded Text Embedding block contains an image encoder and a cross-attention text encoder. We employ a vision transformer as the image encoder to convert visual content into detailed feature representations. The text encoder (e.g., BERT \cite{devlin2019bert} model) leverages a cross-attention mechanism to process textual inputs while interacting with image features. This bidirectional attention mechanism enables image features to focus on relevant text segments while textual representations attend to corresponding image regions, effectively grounding linguistic concepts in their visual counterparts. The image-grounded text encoder processes information through multiple cross-attention layers, generating refined representations where text and image features are deeply contextualized. The output of this vision-language encoder is a set of text embeddings that have been integrated with visual contextual information from the corresponding frame. These embeddings then pass to the GPT-2 backbone for decoding.

\begin{table*}[!ht]
\caption{Comparative analysis of our method against other SOTA surgical VQA solutions on open-ended PitVQA and EndoVis18-VQA datasets. FFT and PEFT denote fully finetuning and parameter-efficient finetuning.}
\label{tab:compare-result}
\centering
\setlength{\tabcolsep}{5.5pt}
\renewcommand{\arraystretch}{1.2}
\begin{tabular}{ccccccccc}
\hline
\multicolumn{9}{c}{\textbf{Open-Ended PitVQA}} \\ \hline
\multicolumn{2}{c|}{Model} & \multicolumn{1}{c|}{BLEU-1(\%)} & \multicolumn{1}{c|}{BLEU-2(\%)} & \multicolumn{1}{c|}{BLEU-3(\%)} & \multicolumn{1}{c|}{BLEU-4(\%)} & \multicolumn{1}{c|}{Rouge-1(\%)} & \multicolumn{1}{c|}{Rouge-L(\%)} & Meteor(\%) \\ \hline
\multicolumn{1}{c|}{\multirow{3}{*}{FFT}} & \multicolumn{1}{c|}{VisualBert \cite{li2019visualbert}} & \multicolumn{1}{c|}{71.11} & \multicolumn{1}{c|}{68.87} & \multicolumn{1}{c|}{66.75} & \multicolumn{1}{c|}{64.02} & \multicolumn{1}{c|}{78.52} & \multicolumn{1}{c|}{78.32} & 76.30 \\
\multicolumn{1}{c|}{} & \multicolumn{1}{c|}{VisualBert RM \cite{seenivasan2022surgical}} & \multicolumn{1}{c|}{75.42} & \multicolumn{1}{c|}{72.09} & \multicolumn{1}{c|}{69.22} & \multicolumn{1}{c|}{65.80} & \multicolumn{1}{c|}{78.36} & \multicolumn{1}{c|}{77.94} & 75.83 \\
\multicolumn{1}{c|}{} & \multicolumn{1}{c|}{PitVQA++ (Ours)} & \multicolumn{1}{c|}{80.28} & \multicolumn{1}{c|}{78.17} & \multicolumn{1}{c|}{76.44} & \multicolumn{1}{c|}{74.11} & \multicolumn{1}{c|}{82.62} & \multicolumn{1}{c|}{82.46} & 82.27 \\ \hline
\multicolumn{1}{c|}{\multirow{4}{*}{\begin{tabular}[c]{@{}c@{}}PitVQA++\\ (PEFT)\end{tabular}}} & \multicolumn{1}{c|}{LoRA \cite{hu2022lora}} & \multicolumn{1}{c|}{80.81} & \multicolumn{1}{c|}{78.45} & \multicolumn{1}{c|}{76.72} & \multicolumn{1}{c|}{74.53} & \multicolumn{1}{c|}{83.24} & \multicolumn{1}{c|}{83.13} & 82.88 \\
\multicolumn{1}{c|}{} & \multicolumn{1}{c|}{MoRA \cite{jiang2024mora}} & \multicolumn{1}{c|}{81.15} & \multicolumn{1}{c|}{78.82} & \multicolumn{1}{c|}{77.10} & \multicolumn{1}{c|}{74.84} & \multicolumn{1}{c|}{83.56} & \multicolumn{1}{c|}{83.47} & 83.36 \\
\multicolumn{1}{c|}{} & \multicolumn{1}{c|}{MoLoRA (Ours)} & \multicolumn{1}{c|}{{\underline{81.63}}} & \multicolumn{1}{c|}{{\underline{79.59}}} & \multicolumn{1}{c|}{{\underline{77.75}}} & \multicolumn{1}{c|}{{\underline{75.48}}} & \multicolumn{1}{c|}{{\underline{84.18}}} & \multicolumn{1}{c|}{{\underline{84.06}}} & {\underline{83.91}} \\
\multicolumn{1}{c|}{} & \multicolumn{1}{c|}{Vector-MoLoRA (Ours)} & \multicolumn{1}{c|}{\textbf{82.24}} & \multicolumn{1}{c|}{\textbf{80.17}} & \multicolumn{1}{c|}{\textbf{78.42}} & \multicolumn{1}{c|}{\textbf{76.17}} & \multicolumn{1}{c|}{\textbf{84.68}} & \multicolumn{1}{c|}{\textbf{84.51}} & \textbf{84.41} \\ \hline
\multicolumn{9}{c}{\textbf{EndoVis18-VQA}} \\ \hline
\multicolumn{2}{c|}{Model} & \multicolumn{1}{c|}{BLEU-1(\%)} & \multicolumn{1}{c|}{BLEU-2(\%)} & \multicolumn{1}{c|}{BLEU-3(\%)} & \multicolumn{1}{c|}{BLEU-4(\%)} & \multicolumn{1}{c|}{Rouge-1(\%)} & \multicolumn{1}{c|}{Rouge-L(\%)} & Meteor(\%) \\ \hline
\multicolumn{1}{c|}{\multirow{3}{*}{FFT}} & \multicolumn{1}{c|}{VisualBert \cite{li2019visualbert}} & \multicolumn{1}{c|}{75.61} & \multicolumn{1}{c|}{72.94} & \multicolumn{1}{c|}{70.16} & \multicolumn{1}{c|}{67.21} & \multicolumn{1}{c|}{80.23} & \multicolumn{1}{c|}{80.17} & 75.95 \\
\multicolumn{1}{c|}{} & \multicolumn{1}{c|}{VisualBert RM \cite{seenivasan2022surgical}} & \multicolumn{1}{c|}{75.98} & \multicolumn{1}{c|}{73.20} & \multicolumn{1}{c|}{70.33} & \multicolumn{1}{c|}{66.40} & \multicolumn{1}{c|}{77.59} & \multicolumn{1}{c|}{77.58} & 76.06 \\
\multicolumn{1}{c|}{} & \multicolumn{1}{c|}{PitVQA++ (Ours)} & \multicolumn{1}{c|}{80.14} & \multicolumn{1}{c|}{79.35} & \multicolumn{1}{c|}{78.44} & \multicolumn{1}{c|}{76.99} & \multicolumn{1}{c|}{88.38} & \multicolumn{1}{c|}{88.42} & 84.86 \\ \hline
\multicolumn{1}{c|}{\multirow{4}{*}{\begin{tabular}[c]{@{}c@{}}PitVQA++\\ (PEFT)\end{tabular}}} & \multicolumn{1}{c|}{LoRA \cite{hu2022lora}} & \multicolumn{1}{c|}{81.33} & \multicolumn{1}{c|}{80.49} & \multicolumn{1}{c|}{79.63} & \multicolumn{1}{c|}{78.29} & \multicolumn{1}{c|}{88.73} & \multicolumn{1}{c|}{88.74} & 85.60 \\
\multicolumn{1}{c|}{} & \multicolumn{1}{c|}{MoRA \cite{jiang2024mora}} & \multicolumn{1}{c|}{\underline{81.68}} & \multicolumn{1}{c|}{80.56} & \multicolumn{1}{c|}{79.28} & \multicolumn{1}{c|}{77.31} & \multicolumn{1}{c|}{\underline{88.99}} & \multicolumn{1}{c|}{\underline{89.01}} & 85.65 \\
\multicolumn{1}{c|}{} & \multicolumn{1}{c|}{MoLoRA (Ours)} & \multicolumn{1}{c|}{{81.55}} & \multicolumn{1}{c|}{{\underline{80.74}}} & \multicolumn{1}{c|}{{\underline{79.91}}} & \multicolumn{1}{c|}{{\underline{78.59}}} & \multicolumn{1}{c|}{{88.89}} & \multicolumn{1}{c|}{{88.91}} & {\underline{85.76}} \\
\multicolumn{1}{c|}{} & \multicolumn{1}{c|}{Vector-MoLoRA (Ours)} & \multicolumn{1}{c|}{\textbf{82.17}} & \multicolumn{1}{c|}{\textbf{81.37}} & \multicolumn{1}{c|}{\textbf{80.51}} & \multicolumn{1}{c|}{\textbf{79.15}} & \multicolumn{1}{c|}{\textbf{89.63}} & \multicolumn{1}{c|}{\textbf{89.64}} & \textbf{86.25} \\ \hline
\end{tabular}
\end{table*}

\textit{GPT-2 Backbone with Vector-MoLoRA:}
The GPT-2 backbone consists of 12 identical transformer blocks, with each block consisting of an attention layer and a feed-forward network. We freeze all transformer blocks and introduce a Vector-MoLoRA adaptor into the combined QKV projection layer (GPT2LMHeadModel.transformer.h[n].attn.c\_attn) of each transformer block. This attention projection layer performs a single linear transformation to generate the Query (Q), Key (K), and Value (V) matrices simultaneously, with dimensions expanding from 768 to 2304. The Vector-MoLoRA adaptor comprises independent LoRA and MoRA modules, as defined in Eq. \ref{eq:vec-molora-1} and \ref{eq:vec_molora_2}. The LoRA module contains two trapezoidal adaptors with shapes $(768,\space r_l^i)$ and $(r_l^i,\space 2304)$, while the MoRA module comprises non-trainable compression and decompression blocks, along with a trainable rectangular adaptor of shape $(r_m^i,\space r_m^i)$. Here, $r_l^i$ and $r_m^i$ are defined in Eq. \ref{eq:lora_vector_2} and \ref{eq:mora_vector}, where $i \in \{1, 2, ..., 12\}$ corresponds to the 12 transformer blocks in GPT-2. Both rank vectors contain monotonically decreasing values that can be customized by users. This design aligns with the hierarchical structure of deep networks and contributes to improved model performance. The compression and decompression operations are defined by Equations \ref{eq:g_func}-\ref{eq:rotation-mat}.
The output representations from the GPT-2 backbone are passed through a language model head and a subsequent softmax layer to produce the final text predictions.

\textit{Automatic Failure Detection:}
To enhance the reliability of AI-driven surgical decision support, PitVQA++ incorporates an entropy-based uncertainty threshold to detect and reject uncertain predictions. During inference, predictions with high entropy values are flagged as unreliable and referred to a clinician for review. This mechanism ensures that ambiguous or low-confidence responses do not influence critical surgical decisions, thereby improving trust, safety, and overall model reliability in open-ended surgical VQA tasks. The risk-coverage curve (e.g., referral curve) in Fig. \ref{fig:risk_curve}, following \cite{band2021benchmarking}, illustrates the impact of threshold selection, demonstrating how the exclusion ratio of uncertain predictions affects the trade-off between expert referrals and acceptable performance.

\section{EXPERIMENTS}
\subsection{Dataset}
In addition to our open-ended PitVQA dataset, we also evaluate our model on a public benchmark dataset of open-ended EndoVis18-VQA \cite{seenivasan2022surgical}. This dataset contains 13,790 question-answer pairs derived from 2,086 surgical scenes across 14 nephrectomy surgery videos.
The questions consist of 6-9 words, while answers contain 5-10 words, with a vocabulary of 50 unique words. We follow the original data split \cite{seenivasan2022surgical} for training and validation sets. Therefore, the training set comprises 1,560 frames and 10,574 question-answer pairs, while the validation set consists of 447 frames and 3,216 question-answer pairs. 

\begin{figure*}[!ht]
    \centering
    \includegraphics[width=0.90\textwidth]
    {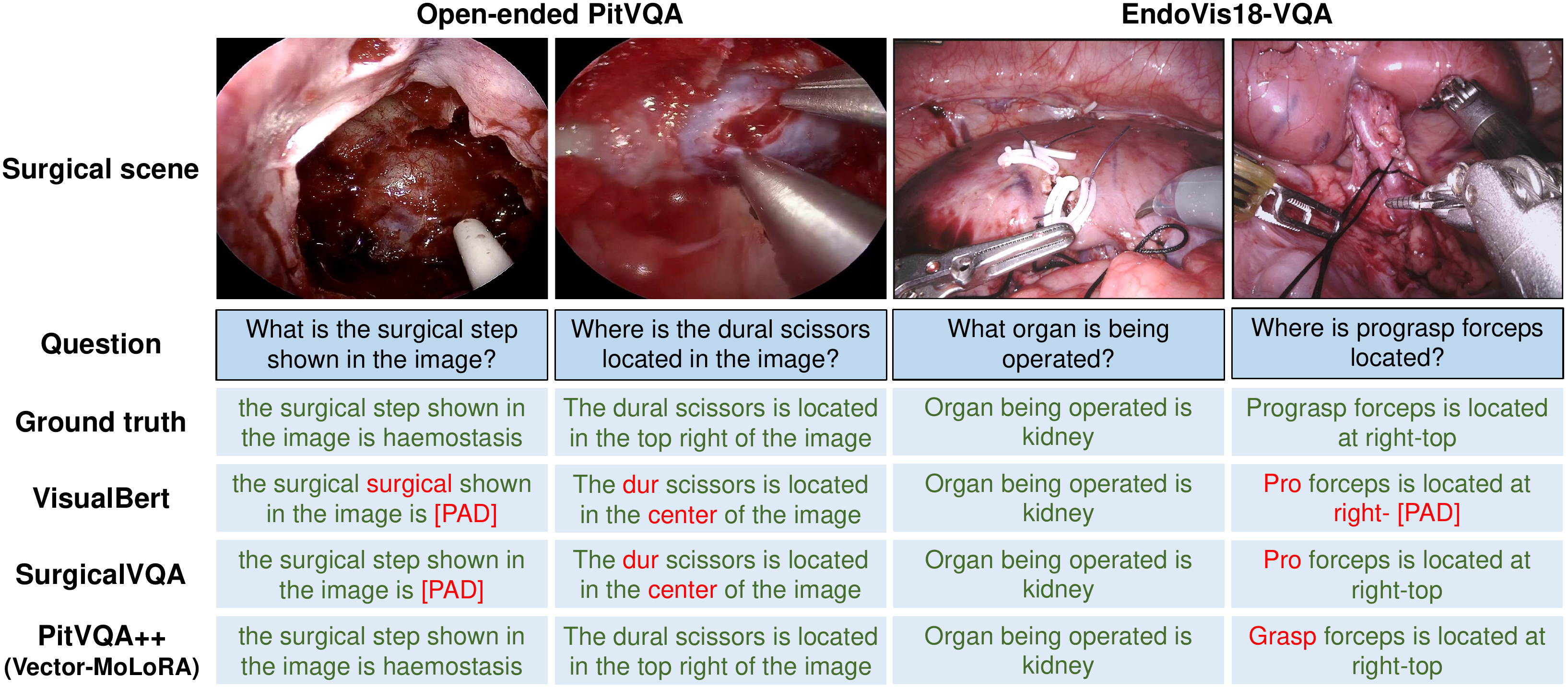}
    \caption{Qualitative comparison of generation results between baseline models and ours on open-ended PitVQA and EndoVis18-VQA datasets.}
    \label{fig:quali_result}
\end{figure*}

\subsection{Implementation Details}
Our backbone networks utilize the implementation and pre-trained weights from the Huggingface official repositories of vision transformer, BLIP, and GPT-2. The model is optimized using Adam optimizer with a learning rate of $2 \times 10^{-7}$ and cross-entropy loss. To evaluate performance, we retrained several state-of-the-art (SOTA) surgical VQA models using their official repositories: VisualBert and VisualBert RM. We report the highest 4-gram Bilingual Evaluation Understudy (BLEU-4) score on the validation set for performance comparison and ablation studies. All experiments are conducted using PyTorch and a single NVIDIA RTX A6000 GPU.

\subsection{Experimental Results}
We conduct comprehensive experiments to validate our proposed method and evaluate the effectiveness of Vector-MoLoRA. We utilize BLEU scores (BLEU-1 through BLEU-4) to assess the lexical accuracy of generated responses, while employing ROUGE metrics (ROUGE-1 and ROUGE-L) and METEOR to measure the semantic coherence between generated and reference answers. These diverse automatic evaluation metrics collectively reflect model's robustness in answer generation across different linguistic dimensions.

Our experimental results and validation studies are presented in table~\ref{tab:compare-result} and Fig.~\ref{fig:quali_result}. We compare our method against state-of-the-art surgical VQA models and parameter-efficient adaptation approaches on Open-ended PitVQA and EndoVis18-VQA datasets, as shown in TABLE~\ref{tab:compare-result}. Compared to the fully fine-tuned VisualBert RM, PitVQA++ with Vector-MoLoRA achieves substantial improvements across all metrics. In the Open-ended PitVQA dataset, it achieves higher scores in BLEU-4, Rouge-L, and Meteor, indicating improved language generation quality. Similar performance gains are observed on the EndoVis18-VQA dataset, underscoring the effectiveness of Vector-MoLoRA in adapting vision language models while maintaining efficiency. Furthermore, when compared to the latest adaptation technique, Vector-MoLoRA outperforms MoRA across BLEU-4, Rouge-L, and Meteor metrics. The improved performance on both the Open-ended PitVQA and EndoVis18-VQA datasets validates that our hierarchical structure design in Vector-MoLoRA not only enhances model adaptation capabilities but also demonstrates strong robustness and generalizability.

We also carry out qualitative analysis on both datasets to visualize model performance, as shown in Fig.~\ref{fig:quali_result}. We observe that while previous SOTA models accurately generate common terms (e.g., kidney, scissors) from both datasets, our model excels at generating surgical terminology (e.g., haemostasis) on the Open-ended PitVQA dataset. We observe on both datasets that VisualBert and VisualBert RM frequently generate word fragments (e.g., 'dur' instead of 'dural'), whereas our method successfully generates complete terms. This superior generation capability stems from PitVQA++'s enhanced learning through its cross-attention-based multimodal fusion and the hierarchical rank allocation strategy implemented in Vector-MoLoRA. Additionally, PitVQA++ demonstrates superior performance in visual reasoning. For example, as shown in the second column of Fig. ~\ref{fig:quali_result}, our method successfully identifies the dural scissors in the top right corner, while previous solutions fail to distinguish between different instruments.

\subsection{Ablation Studies}
We evaluate our method through three ablation studies: (1) experimental verification of Vector-MoLoRA's effectiveness in mitigating catastrophic forgetting, (2) analysis of vector configurations and their performance impact, and (3) reliability analysis for Vector-MoLoRA using risk coverage curves.

\begin{table}[!ht]
\caption{Ablation Study on Mitigation of Catastrophic Forgetting}
\label{tab:ablation-1}
\centering
\setlength{\tabcolsep}{4pt}
\renewcommand{\arraystretch}{1.15}  
\begin{tabular}{c|c|cc|cc}
\hline
\multirow{2}{*}{\textbf{\begin{tabular}[c]{@{}c@{}}Pretrained \\ weights / \\ Training data\end{tabular}}} & \multirow{2}{*}{\textbf{\begin{tabular}[c]{@{}c@{}}PitVQA++\\ Network\end{tabular}}} & \multicolumn{2}{c|}{\textbf{\begin{tabular}[c]{@{}c@{}}Training Domain\\ Validation\end{tabular}}} & \multicolumn{2}{c}{\textbf{\begin{tabular}[c]{@{}c@{}}Pretrained Domain\\ Validation\end{tabular}}} \\ \cline{3-6} 
 &  & \multicolumn{1}{c|}{\begin{tabular}[c]{@{}c@{}}Rouge-L\\ (\%)\end{tabular}} & \begin{tabular}[c]{@{}c@{}}Meteor\\ (\%)\end{tabular} & \multicolumn{1}{c|}{\begin{tabular}[c]{@{}c@{}}Rouge-L\\ (\%)\end{tabular}} & \begin{tabular}[c]{@{}c@{}}Meteor\\ (\%)\end{tabular} \\ \hline
\begin{tabular}[c]{@{}c@{}}GPT2 / \\ Open-ended \\ PitVQA\end{tabular} & FFT & \multicolumn{1}{c|}{80.21} & 80.02 & \multicolumn{1}{c|}{--} & -- \\ \hline
\begin{tabular}[c]{@{}c@{}}Open-ended \\ PitVQA / \\ EndoVis18-\\ VQA\end{tabular} & FFT & \multicolumn{1}{c|}{90.95} & 88.71 & \multicolumn{1}{c|}{52.68} & 51.13 \\ \hline
\begin{tabular}[c]{@{}c@{}}Open-ended \\ PitVQA / \\ EndoVis18-\\ VQA\end{tabular} & \begin{tabular}[c]{@{}c@{}}Vector-\\ MoLoRA\end{tabular} & \multicolumn{1}{c|}{91.43} & 89.74 & \multicolumn{1}{c|}{75.90} & 75.21 \\ \hline
\end{tabular}
\end{table}

\subsubsection{Mitigation of Catastrophic Forgetting}
Table~\ref{tab:ablation-1} shows the effectiveness of Vector-MoLoRA in mitigating catastrophic forgetting during transfer learning from open-ended PitVQA to EndoVis18-VQA. First, we fine-tune the PitVQA++ backbone network (i.e., multimodal encoder and GPT-2 decoder) on open-ended PitVQA dataset with frozen Vector-MoLoRA adapters. Using these pre-trained weights, we then conduct two fine-tuning strategies on EndoVis18-VQA: 1) Full Fine-tuning of the PitVQA++ backbone network, and 2) Vector-MoLoRA adaptation with a frozen GPT-2 decoder. As shown in the results, PitVQA++ with Vector-MoLoRA achieves superior performance on the pretrained domain validation while maintaining comparable performance on the training domain validation compared to its fully fine-tuned version. This demonstrates Vector-MoLoRA's effectiveness in mitigating catastrophic forgetting during transfer learning.

\begin{table}[!ht]
\caption{Ablation study of different rank Vectors on open-ended PitVQA dataset.}
\label{tab:ablation-2}
\centering
\renewcommand{\arraystretch}{1.2}
\setlength{\tabcolsep}{4pt}
\begin{tabular}{ccccc}
\hline
\multicolumn{1}{c|}{\begin{tabular}[c]{@{}c@{}}BLEU-3\\ (\%)\end{tabular}} & \multicolumn{1}{c|}{\begin{tabular}[c]{@{}c@{}}BLEU-4\\ (\%)\end{tabular}} & \multicolumn{1}{c|}{\begin{tabular}[c]{@{}c@{}}Rouge-L\\ (\%)\end{tabular}} & \multicolumn{1}{c|}{\begin{tabular}[c]{@{}c@{}}Meteor\\ (\%)\end{tabular}} & \begin{tabular}[c]{@{}c@{}}Rank Vectors\\ $(r_{1}, ..., r_{12})$\end{tabular} \\ \hline
\multicolumn{1}{c|}{77.67} & \multicolumn{1}{c|}{75.42} & \multicolumn{1}{c|}{84.46} & \multicolumn{1}{c|}{84.01} & \begin{tabular}[c]{@{}c@{}}MoRA: (80, ..., 40)\\ LoRA: (40, ..., 20)\end{tabular} \\ \hline
\multicolumn{1}{c|}{78.16} & \multicolumn{1}{c|}{75.91} & \multicolumn{1}{c|}{\underline{84.59}} & \multicolumn{1}{c|}{\underline{84.33}} & \begin{tabular}[c]{@{}c@{}}MoRA: (72, ..., 32)\\ LoRA: (36, ..., 16)\end{tabular} \\ \hline
\multicolumn{1}{c|}{\textbf{78.42}} & \multicolumn{1}{c|}{\textbf{76.17}} & \multicolumn{1}{c|}{84.51} & \multicolumn{1}{c|}{\textbf{84.41}} & \begin{tabular}[c]{@{}c@{}}MoRA: (64, ..., 24)\\ LoRA: (32, ..., 12)\end{tabular} \\ \hline
\multicolumn{1}{c|}{\underline{78.29}} & \multicolumn{1}{c|}{\underline{76.05}} & \multicolumn{1}{c|}{\textbf{84.69}} & \multicolumn{1}{c|}{\textbf{84.41}} & \begin{tabular}[c]{@{}c@{}}MoRA: (56, ..., 16)\\ LoRA: (28, ..., 8)\end{tabular} \\ \hline
\multicolumn{1}{c|}{78.10} & \multicolumn{1}{c|}{75.83} & \multicolumn{1}{c|}{84.36} & \multicolumn{1}{c|}{84.23} & \begin{tabular}[c]{@{}c@{}}MoRA: (56, ..., 16)\\ LoRA: (18, ..., 8)\end{tabular} \\ \hline
\multicolumn{1}{c|}{77.74} & \multicolumn{1}{c|}{75.45} & \multicolumn{1}{c|}{84.10} & \multicolumn{1}{c|}{84.00} & \begin{tabular}[c]{@{}c@{}}MoRA: (36, ..., 16)\\ LoRA: (28, ..., 8)\end{tabular} \\ \hline
\multicolumn{1}{c|}{77.35} & \multicolumn{1}{c|}{75.05} & \multicolumn{1}{c|}{83.88} & \multicolumn{1}{c|}{83.65} & \begin{tabular}[c]{@{}c@{}}MoRA: (18, ..., 8)\\ LoRA: (18, ..., 8)\end{tabular} \\ \hline
\end{tabular}
\end{table}

\subsubsection{Analysis of Different Vector Configurations}
Table~\ref{tab:ablation-2} presents the performance of PitVQA++ with Vector-MoLoRA across different vector configurations on the Open-Ended PitVQA dataset, with each row representing a distinct configuration. The third configuration achieves the best performance, where MoRA and LoRA vectors are decremented every two layers, with MoRA decreasing by 8 and LoRA decreasing by 4. The first, second, and fourth configurations maintain the same step length but start with different MoRA and LoRA rank values. The last three configurations use smaller step sizes and rank values. The results show that the model performance improves as the rank values of MoRA and LoRA vectors increase, reaching its peak with MoRA ranks decreasing from 64 to 24 and LoRA ranks from 32 to 12. However, further increasing the rank values leads to performance degradation, as larger ranks introduce more parameters, making the model prone to overfitting.

\begin{figure}[!ht]
    \centering
    \includegraphics[width=1\columnwidth]
    {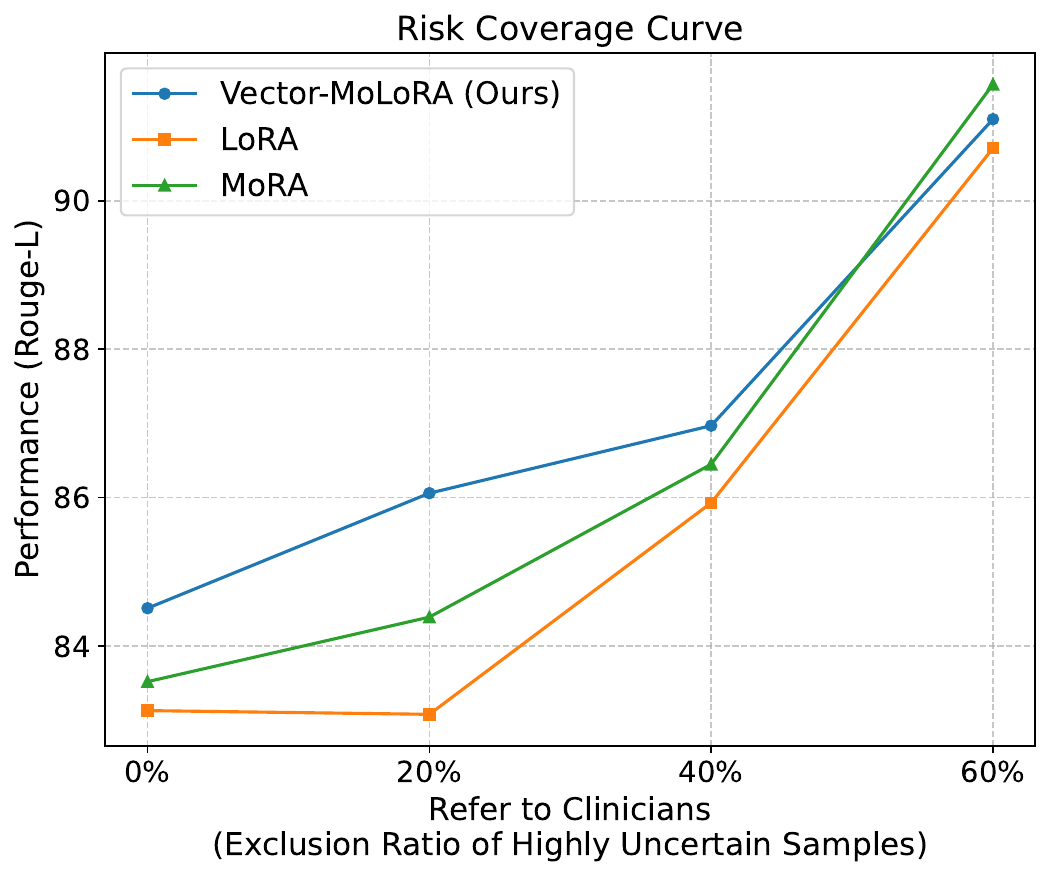}
    \caption{Risk coverage curves of PitVQA++ with LoRA, MoRA, and Vector-MoLoRA (Ours) on open-ended PitVQA dataset.}
    \label{fig:risk_curve}
\end{figure}

\subsubsection{Analysis of Reliability}
The risk-coverage analysis in Fig.~\ref{fig:risk_curve} demonstrates that our proposed Vector-MoLoRA exhibits greater reliability than LoRA and MoRA by achieving higher performance (ROUGE-L) while requiring fewer expert referrals for uncertain predictions. For example, when the 20\% most uncertain samples are excluded during inference, our method achieves a 1.55\% ROUGE-L improvement, whereas LoRA and MoRA achieve -0.06\% and 0.87\%, respectively.
This highlights Vector-MoLoRA’s superior confidence calibration, ensuring that uncertain predictions are handled more effectively. By referring only the most ambiguous cases to clinicians, our approach enhances trust in AI-driven decision-making, optimizing intraoperative guidance and supporting more efficient postoperative training programs.


\section{Discussion and Conclusions}
In this study, we introduced PitVQA++, a novel Vision-Language Model (VLM) tailored for open-ended Visual Question Answering (VQA) in endonasal pituitary adenoma surgery. Our approach integrates Vector-MoLoRA, an innovative parameter-efficient adaptation strategy that dynamically allocates low-rank and matrix-rank updates across transformer layers based on hierarchical feature importance. In addition, we presented Open-Ended PitVQA, a comprehensive surgical VQA dataset that captures intraoperative decision-making, procedural steps, and surgical tool interactions, significantly expanding the scope of open-ended question-answering in surgical AI.

Experimental evaluations on both Open-Ended PitVQA and the EndoVis18-VQA dataset demonstrated that PitVQA++ outperforms existing state-of-the-art surgical VQA models, particularly in generating more contextually relevant and linguistically precise responses. The ablation studies provided further insights into the efficacy of Vector-MoLoRA, validating its ability to mitigate catastrophic forgetting while enhancing fine-tuning efficiency. Our risk-coverage analysis further confirmed that Vector-MoLoRA achieves higher reliability by improving performance under uncertainty, requiring fewer expert referrals for ambiguous predictions compared to LoRA and MoRA, thus enhancing its applicability for real-time clinical decision support.

These findings highlight the potential of PitVQA++ as a trustworthy AI assistant for intraoperative decision-making, offering surgeons context-aware, interpretable, and highly reliable responses. Future research directions include expanding the dataset to incorporate multimodal surgical cues (e.g., audio and haptic feedback), refining uncertainty quantification techniques for improved trust calibration, and exploring more advanced vision-language fusion mechanisms to further enhance system robustness and usability in real-world surgical settings.

\balance
\bibliographystyle{IEEEtran}
\bibliography{mybib}

\end{document}